\documentclass{ieeeaccess} 

\usepackage{cite} 

\usepackage{amsmath,amssymb,amsfonts} 

\usepackage{algorithmic} 

\usepackage{graphicx} 

\usepackage{textcomp} 

\usepackage{tabularx,,booktabs} 

\usepackage{dblfloatfix} 

\usepackage{float} 

\usepackage{nidanfloat} 

\usepackage{url} 

\usepackage{multirow}

\def\BibTeX{{\rm B\kern-.05em{\sc i\kern-.025em b}\kern-.08em 

    T\kern-.1667em\lower.7ex\hbox{E}\kern-.125emX}} 

\begin{document} 

\history{Date of publication xxxx 00, 0000, date of current version xxxx 00, 0000.} 

\doi{---}

\title{District Wise Price Forecasting of Wheat in Pakistan using Deep Learning} 

\author{\uppercase{Ahmed Rasheed}\authorrefmark{1},  \uppercase{Muhammad Shahzad Younis}\authorrefmark{1}, \uppercase{Farooq Ahmad}\authorrefmark{1}, \uppercase{Junaid Qadir\authorrefmark{2}, and Muhammad Kashif}.\authorrefmark{1}} 
\address[1]{National University of Sciences and Technology, Islamabad, Pakistan.} 
\address[2]{ Information  Technology  University (ITU), Lahore, Pakistan.  
} 





\corresp{Corresponding author: Ahmed Rasheed (e-mail: arasheed.msee17seecs@seecs.edu.pk).}

\begin{abstract} 

Wheat is the main agricultural crop of Pakistan and is a staple food requirement of almost every Pakistani household making it the main strategic commodity of the country whose availability and affordability is the government's main priority. Wheat food availability can be vastly affected by multiple factors included but not limited to the production, consumption, financial crisis, inflation, or volatile market. The government ensures food security by particular policy and monitory arrangements, which keeps up purchase parity for the poor. Such arrangements can be made more effective if a dynamic analysis is carried out to estimate the future yield based on certain current factors. Future planning of commodity pricing is achievable by forecasting their future price anticipated by the current circumstances.  This paper presents a wheat price forecasting methodology, which uses the price, weather, production, and consumption trends for wheat prices taken over the past few years and analyzes them with the help of advance neural networks architecture Long Short Term Memory (LSTM) networks. The proposed methodology presented significantly improved results versus other conventional machine learning and statistical time series analysis methods.

\end{abstract}


\titlepgskip=-15pt

\maketitle

\section{Introduction} 

\PARstart{P}{akistan} is an agrarian economy in which a major of livelihoods are intimately connected with the agriculture sector. Wheat is the main agricultural crop of Pakistan. It is estimated that 80\% of Pakistani farmers grow wheat on an area of around nine million hectares (close to 40\% of the country's total cultivated land) \cite{b32}. Wheat heavily influences Pakistan's gross domestic product (GDP), and it also adds to the earnings brought by foreign exchange. Specifically, wheat accounts 8.9\% value-added in agriculture and 1.6\% of GDP according to them according to the Government of Pakistan (GOP) statistics \cite{b33}. Wheat, in particular, is a crop that provides linkages that stimulate the growth of economics in other sectors. In Pakistan, wheat is also being cultivated in a group shaped cropping system including wheat-cotton, wheat-rice, wheat-sugarcane, and wheat-maize. Of the total areas where the area cultivation is dependent on rain irrigation (an area of roughly 1.5 million hectares), wheat-cotton and wheat-rice cropping constitute approximately 60\%\cite{b34}. The percentage of the area which is being used for the cultivation of agricultural purposes is increasing to a relatively higher degree \cite{c1}.


However, the emphasis on the importance of agriculture has been decreasing over time. The main reason for it can be attributed to the increased urbanization of the population. Even then, there is still a large percentage of the population that lives in rural areas and indulges in the agricultural sector.  Various problems have been affecting the cultivation of wheat. Some of these problems include the shortage of water for irrigation, shortage and increased prices of key input, and outdated traditional farming methods. 


Due to these problems, Pakistan has experienced various ups and downs in the case of the production of wheat. During the seasons in which production is low, the prices of wheat flour and raw wheat boosts up, but it also decreases where there is an increased production of wheat. However, the surplus amount of wheat production affects the farming community, which suffers from a significant loss in revenue. This is mainly because of the insufficient amount of marketing facilities. Therefore, the farmers keep in mind the prospects of future production and prices at the time of cultivation. Hence, there is a need to forecast the area of the cultivation, the yield, and also the production. Such analysis can help in keeping a healthy balance between pricing and production. Hence, this paper will focus on determining the prospects of wheat in Pakistan by using past trends as its base.

Future market commodity pricing is of great interest to governments, investors, and producers. Food type commodities are traded in smaller localized markets, which constitute gross effect towards its supply and pricing. Commodities like wheat have a high impact on the local economy and food security for the poor. Such commodities are highly traded and are prone to price fluctuation because of multiple factors, including a higher flux of investment, gross production, global financial outlook, and monitory policies. In the presence of such complex and highly dynamic factors, commodity pricing forecasting is needed to consider all impactful trends to provide a reliable resource to investors and governments to reduce risks related to price volatility. Forecasting is the primary tool used again for the prediction on the base of the existing and past knowledge base. With the forecasting, one can predict the production of wheat and determine the parameters that will influence production in the coming years. This is done mainly to gather the interest of the people and to attract them to grow the main selling crops more as compared to others and also to attract people who are not in farming already.

In Pakistan, the forecast for the production of major crops is carried out by the agriculture department. Forecasting is mainly done to find out an estimate of the supply and demand of the produce. It is issued publicly by the United States Department of Agriculture (USDA, 1999) and is used not only in the United States but also all throughout the world, including Pakistan. However, in Pakistan, the forecast is only made for the production of the crop. Forecasting of wheat's supply and demand is foreseen as there is no specialized mechanism designed for the prices of the crops. Another reason is the uncertainties associated with the behavior of the buyers and sellers as the demand and supply fluctuate. During 2014, the support price of wheat was Rs. 1200/40kg, as announced by the government. This may have a significant effect on the future cropping pattern. The wheat cultivation has increased significantly from the years 2011 to 2013 from 8.650 million hectares in 2011 to 8.693 million hectares in 2013. This increment in crop cultivation is mainly because of a gradual rise in the prices of the crop, which was about Rs. 200, which can be attributed to the increase in the size of the grain, which was because of the proper irrigation caused by the timely rains, favorable temperature, and good quality seeds.

Our main contribution in this study is the forecasting of the trends found in the prices over time and the testing of these forecasts with the future trends of the wheat crop in Pakistan. This will be done by using appropriate trend sensitive prediction models to forecast the prices. In this study, we have formulated two cases to predict the future price values of wheat: in the first one, the price is predicted for only the next upcoming month, while in the latter,  we have forecasted prices of the next whole year. The latter case is a unique method by using which the policy and decision-makers can work efficiently by having a sizeable forecast in the future. This forecasting and the trend analysis will help the future policymakers of the country's agricultural department to make necessary policy decisions also the farmers and market brokers to base their decisions on future market outlook. 

\label{sec:introduction}


\section{Related work}

Much work has been done on the forecasting of stock and commodity prices using the time series analysis techniques while using deep learning models to predict and forecast prices are relatively new. Prices for crops, including wheat, are a highly fluctuating quantity and therefore require prediction methods that can provide better accuracy compared to past practices. 

As mentioned in an article by Friedman et al. \cite{b25} about the impact of the food crisis and domestic wheat prices in Pakistan. Between the years 2005 to 2008, those prices rose by about 106\%. This increase in prices was not only the impact of the global food crisis, but it also depended on the household hoarding and the smuggling of wheat produced. He also mentioned how this situation was taken under control. In 2009, the attainment price of wheat was increased, so the domestic prices were stabilized. According to Ahsan et al. (2011) \cite{b26}, the decline in the quantity of produced wheat is usually related to the increase in prices. In Pakistan, this is mainly because of the natural disasters that may cause water shortage. Another reason is the increased prices of food at an international level because of the increasing economic growth in known countries like China and India. These references make it imperative for prediction systems to be evolved to avoid a last-minute crisis in a country still under development.

Forecasts that are reliable and timely are essential for providing valuable and significant inputs which are used to get help in planning the agriculture and cultivation, even though they are ordinarily full of uncertainties. In this past research \cite{e1}, we find Rani and Raza conducted a study on the forecasting of the price of pulses in Pakistan. They used the Double Exponential Smoothing Function (DESF) as well as the measures of accuracy such as Mean Absolute Percentage Error (MAPE) and Root Mean Squared Error (RMSE) to gain the values of the forecasting. Data from the years 1975 to 2010 were collected based on average yearly prices gathered from various issues of the Agriculture Statistics of Pakistan.

When we look at the previously published researches, it has been proposed that the upcoming prices and the price projection can be guessed by using base forecasts. Moreover, they can also be used to evaluate pre-variation chances. Basis comparison was established for the practical methods of forecasting for raw wheat and wheat flour \cite{d1}.

Machine learning has been used for the prediction of prices for several quantities. It has also been compared to statistical models in terms of accuracy. Juntao Wang (2018) \cite{b31} used Gaussian Process Kernels, a statistical machine learning model, to predict prices of housing. The research demonstrated that Gaussian kernels performed better in terms of price predictions compared to other kernels, such as exponential or quadratic.

Amin Azari (2019) used the Auto-regressive Integrated Moving Average (ARIMA) model in his research on price prediction for the vastly popular and immensely valuable currency, bitcoin \cite{b27} The study also states that the ARIMA model proves to be insufficient to capture the volatility and sharp fluctuations of price.

There are various kinds of neural networks. For time series analysis, the most commonly used architecture is Recurrent Neural Network (RNN) \cite{b20}. These network architectures perform well in retaining the pattern of the data given to them by having a recurrent connection with unit delays in them. In RNNs, the feedback path also exists between the layers, which allow the network also to learn the specific portion of the pattern given to it \cite{a9,a10}.

Long Short Term Memory (LSTM) is a particular type of RNN which perform better by having a memory cell in it [2]. LSTMs are widely used recently to predict the stocks \cite{b35}, weather data \cite{b36}, languages \cite{b37} and many other useful future predictions.  

Karakoyun et al. (2018) \cite{b28} demonstrated a clear difference in results when ARIMA is compared with deep learning (LSTM) for bitcoin value prediction for the next 30 days. Both ARIMA and LSTM are widely used for time-series predictions. However, LSTM is more efficient in evaluation.

Similarly, another comparison between ARIMA and LSTM was discussed by Akbar Siami Namini (2018) \cite{b29}, where both methods were used for the forecasting of economic and financial time series. The purpose of the research was to discuss the positions of deep learning models as compared to traditional statistical models. The research also resulted in the superiority of LSTM over ARIMA.

The survey by Kamilaris et al. (2018)\cite{b30} states the use of deep learning models in agricultural fields. While deep learning has been widely used in agriculture for areas such as plant recognition, crop classification, and classification of land covers. Deep learning, on its own, is also widely used for cost forecasting; however, it is seldom that the two aspects of agriculture and cost forecasting are combined. The survey also states that a vast majority of the research done in this regard dates to 2015 or onward, which indicates that deep learning is a fairly new direction in agriculture. 






In this study, we are extending the work done by these previous works and providing a comparison of deep learning models with statistical models used for time series analysis catered specifically to crop-price prediction.

Saqib Shakeel et al. \cite{c1} proposed a non-linear model fitting and trend analysis techniques, which include exponential, quadratic, and S-curve models to predict the upcoming prices of wheat and rice in Pakistan. The data used in this research was yearly data from 1975 to 2013, which is quite limited. Extending his work, we are proposing a neural network approach to get better predictions for the price of wheat every month.

\section{Prediction Techniques}

There are several techniques used to get time series prediction. A few of the popular techniques used in recent studies for predicting the prices are used by us to compare the results of our proposed model for prediction.

\subsection{Bagging Trees} 

Bagging Trees are the ensembled form of decision trees, which is a tree-like representation of the possible decisions and their outcomes. Decision trees have had multiple uses in the investment and financing world for just about anything from gold returns \cite{a23} to corporate financial distress \cite{b24} and measurement of firm performances \cite{a25}. 


Bagging or Bootstrap Aggregation is used when the total variance of a decision tree needs reduction. The concept is to generate multiple subsets from training data at random with replacement, which results in the training of multiple decision trees using different subsets. The average of all the results from the decision trees is taken which makes up for a more robust computation than one single decision tree \cite{a4,a5,a6}.

\subsection{Gaussian Process Regression (GPR)} 

GPR is an implementation of the Gaussian Process for regression purposes. GPR provides an ability to compute uncertainty measures for prediction functions using a Bayesian approach to the regression that is non-parametric. 

Since a functional form does not limit GPR, it calculates the probability distribution over all the functions that fit the dataset, unlike Gaussian distribution that calculates the probabilities of a set specific function. A prior is specifies, and the posterior is calculated using training data, the predictive posterior distribution is then computed on specific points of interest \cite{a7}. The underlying representation of GPR is given below.

\begin{equation} 
    y=x^T \beta+\epsilon 
\end{equation}

Here $\epsilon~(0,\sigma_{2})$ is a Gaussian distribution with mean at 0. The error variance $\sigma_{2}$ and coefficients of  $\beta$ are optimally calculated respective to data.

\subsection{Long Short Term Memory Network} 

Neural networks work similarly to our body's neural system. The small neurons are linked to each other to form a single compound structure. 

LSTM network is a modified form of a recurrent neural network to perform better \cite{a1}. The working of the RNN model is shown in Figure \ref{fig1}.


\Figure[b][width=12cm]{rnn.png} 
{Recurrent Neural Network. Here \textit{x, y} and \textit{W} represents the input, output and weights while \textit{a} represents the previous state of the network.\label{fig1}}

RNNs suffer from the problem of vanishing gradient as it propagates back between the layers, so in order to resolve the problem of vanishing gradient, LSTM works on the principle of gates, which allow the network to choose the weights to keep or to discard while training. Three gates, input gate, output gate, and forget gate constitute an LSTM cell \cite{a2}. The input gate is responsible for providing new inputs to the cell. The output gate specifies the output of the cell, and forget the gate is responsible for indicating any prior values that might be needed in the future and retains them. An LSTM-NN cell structure is shown in Figure \ref{rnn_cell}.

\Figure[h]()[width=8cm]{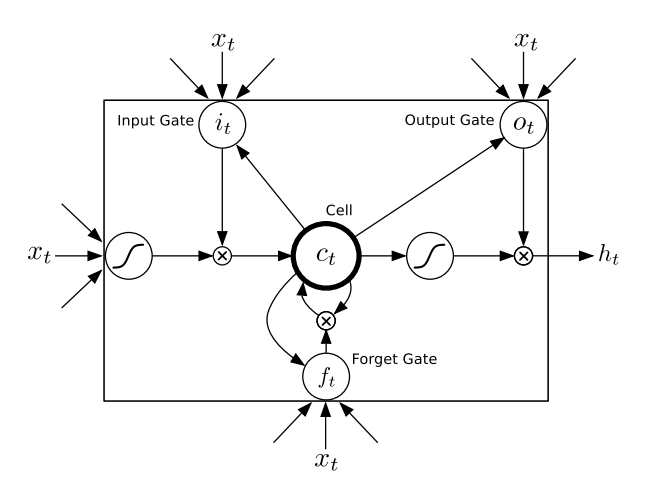} 
{Basic structure of a LSTM-NN cell \cite{b23}.\label{rnn_cell}}

LSTM cells are stacked with each other to form a dence network followed by a fully connected dense layer. The LSTM used in the proposed study comprises of 50 LSTM cells to give the prediction. Hyper-parameters used for the training purpose of LSTM are shown in Table \ref{hyper}.

\begin{table}[h] 

\centering 

\caption{Network Hyper-parameters} 

\label{table}

\begin{tabular}{|c|c|} 

\hline 

\textbf{Parameters}& \textbf{Values} \\ 

\hline  

Max Epochs & 50 \\ 

Mini Batch Size&10\\ 

Loss&MAE\\ 

Activation Function&ReLu\\ 

Optimizer&ADAM\\ 

\hline 

\end{tabular} 

\label{hyper} 

\end{table}

\section{Data}

Monthly prices of wheat for three major districts of Pakistan which are Faisalabad, Gujaranwala and Multan from year 1991 to 2018 were obtained from the Agriculture Marketing Information Service (AMIS) Pakistan using their online Application Program Interfaces (APIs) \cite{b22}. Data on prices for these three districts is used for this research. Figure \ref{fig_orig} shows the price trend over the period with a sampling time of one day.







Smoothing filter was applied to clear out noise and irregularities to help out forecasting algorithms to understand the underlying trend in data better. Such approaches are helpful if a predictor set has different granularity among variables. The smoothing function used in this research is the moving average filter; it takes a defined window of inputs and takes the average of all those values and then applies it to the selected one. 







Along with the price dataset, to aid the networks in forecasting values more efficiently, we also gathered data that shows a correlation with the prices of wheat within the same time intervals. Table \ref{features} states the features picked to aid the networks in forecasting prices along with their co-relation with the prices. The data with yearly granularity was mapped to the monthly by linearly interpolating the values between the two years.

\begin{table}[!htb] 

\centering 

\caption{List of Features used to aid the prediction of prices}

\begin{tabular}{|l|l|l|c|} 

\hline 

\multicolumn{1}{|c|}{\textbf{Sr.}} & \multicolumn{1}{c|}{\textbf{Features}} & \multicolumn{1}{c|}{\textbf{Granularity}} & \multicolumn{1}{c|}{\textbf{\begin{tabular}[c]{@{}c@{}}Co-relation\\ with prices\end{tabular}}} \\ \hline 

1 & Domestic consumption of wheat & Monthly Data & 0.93 \\ 

2 & Growth rate of wheat & Yearly data & 0.67 \\ 

3 & Production of wheat for province & Yearly Data & 0.89 \\ 

4 & Production of wheat for Pakistan & Yearly Data & 0.92 \\ 

5 & Rainfall & Monthly Data & 0.12 \\ 

6 & Average Temperature & Monthly Data & 0.02 \\ \hline

\end{tabular} 

\label{features} 

\end{table}

The Rainfall and Temperature data introduces seasonality to prices of wheat because it has a regular pattern and also predicts the small changes in the prices as it mimics the present time of year and also tells about the effect of severity of the weather on the wheat prices. The yearly data is interpolated on every month to match the size of other features and the output.

Data for domestic consumption, growth rate, and production of wheat were also obtained from AMIS Pakistan. Rainfall and temperature data is obtained from the World Bank Climate Change Knowledge Portal \cite{b21}. All of this collected data act as input features to our model under study.

The data obtained is further divided into two subsets, training data, and test data. The training dataset consists of data from the year 1991 to 2015 and used by the models to train themselves to perform predictions. The test dataset consists of the last three years from the dataset, 2016 to 2018, which is treated as unseen data to the models, and once the model is fully trained, we checked its performance for the prediction of prices for unseen data period.

\section{Methodology}


\Figure[!t]()[width=12cm]{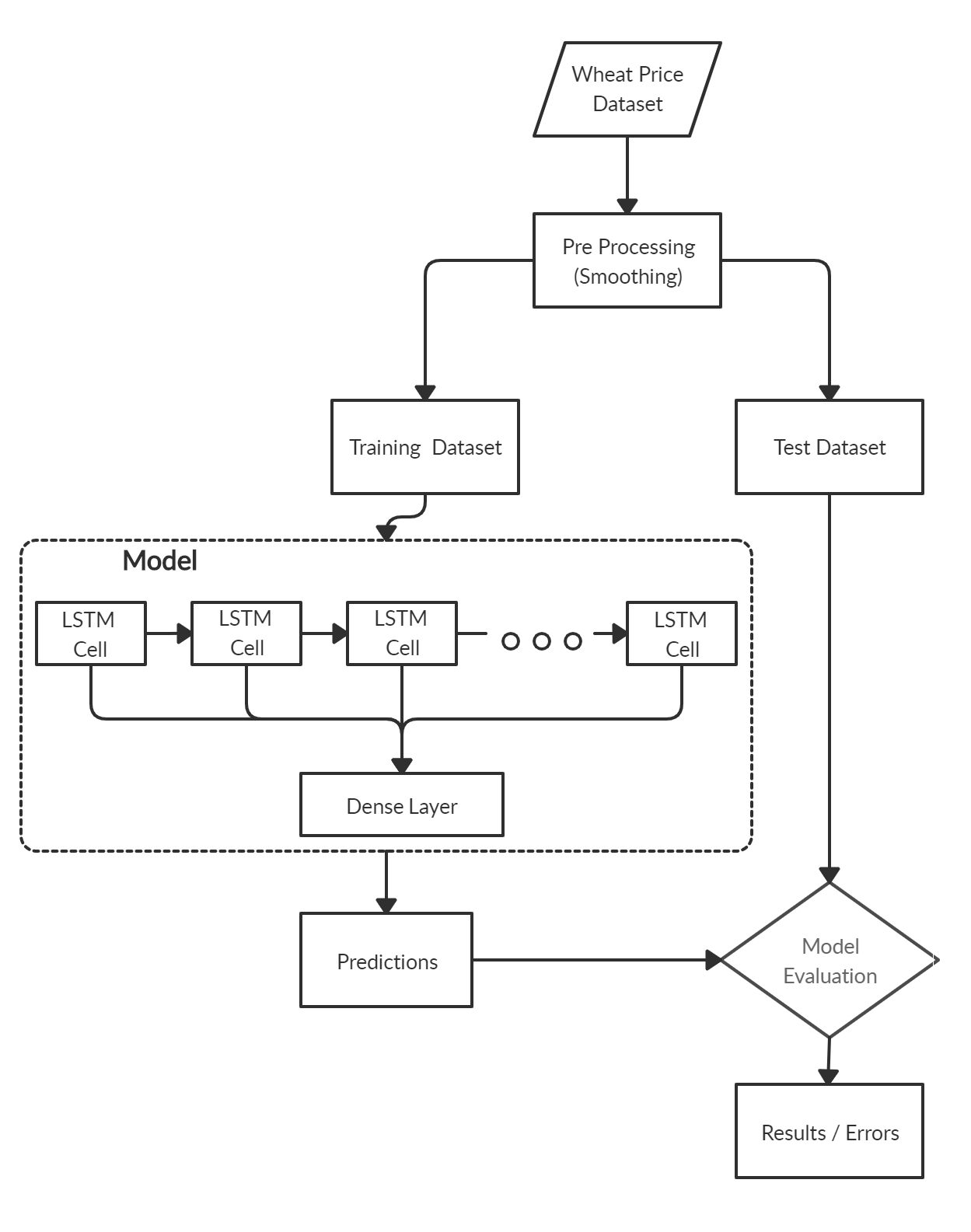} 
{Flow diagram of proposed system.\label{method}} 


Research is initiated with the acquisition of a wheat price dataset. The acquired dataset is split into training and test datasets, whereas training data is further processed before feeding it into the machine and deep learning algorithms. Trained algorithm models are evaluated with training data and error performance (MAPE). 

Metrics are calculated on the basis of which best model is discussed for both processed and raw data. Figure \ref{method} shows the basic flow of the work we have done in the proposed study.

\subsection*{Preprocessing via MA Filter}

Preprocessing refers to the usage of methods and computation to shape raw data into a processable form. The method used for this research is the Moving Average Filter (MA Filter). The filter is a simple low-pass filter that takes an average of multiple samples and gives a single output point. This results in the smoothing of the data array, which is much more feasible and easy for use with neural networks \cite{a8}. We set the window size of 10 values, the mean of these 10 values will be replaced by the actual value in the dataset and eventually the whole dataset is smoothed out. In the Equation \ref{mov_avg} below, the formula of moving average filter is given. Here \textit{x} and \textit{y} are the input and output signals respectively, while \textit{W} is the size of the window used for averaging out the input signal. The original and smoothed data trend is shown in Figure \ref{fig_orig}. 


\begin{equation} 
y[i]=\frac{1}{W} \sum_{j=0}^{W-1} x[i+j] 
\label{mov_avg} 
\end{equation}

\Figure[t]()[width=12cm]{Picture11.PNG} 
{Original and smoothed wheat prices of (a) Faisalabad, (b) Gujranwala and (c) Multan from 1991 to 2018.\label{fig_orig}}



The tests were performed by using both the raw and smoothed data. Since the raw data consists of varying fluctuation, which might affect the results of the forecasting, we used a moving average filter to smooth the input data. This moving filter was used because it helps in not losing essential data in the smoothing process. Figure \ref{method} shows how the data is split into training and test data before applying the smoothing process. It also represents the training and testing on both the datasets using above mentioned LSTM model which brings the best possible results.

Further, we divided the tests into four unique yet similar cases, which will be predicting wheat prices. These cases are explained next.

\subsection*{Case 1: One-month prediction with price data}

In case 1, we predicted the price of wheat only for the upcoming month based on the previous values of the costs, the number of prior values used was varied to find the best results. This forms a univariate system where we are using only one feature to predict a single value as an output.

\subsection*{Case 2: One-month prediction with all features}

In case 2, we used a similar architecture, but this time used all the features of the dataset, along with previous price data, to predict the price of wheat for the next month. These multiple numbers of features help the system to predict the values more efficiently. The list of features used alongside the previous price values is shown in Table \ref{features}

\subsection*{Case 3: One-year prediction with price data} 

In case 3, instead of forecasting a singular upcoming value, we have predicted the prices for a whole year. This brings forth the price forecasting of 12 months by only feeding the previous price values of a few months.

\subsection*{Case 4: One-year prediction with all features} 

In case 4, we predicted the prices for a whole month while feeding all the available features to the models. This method made the network a little complex, but it can be overlooked as it helped in getting good forecast results for the price of the upcoming year. This forms a multivariate system, which takes the number of values as described in Table \ref{features} as input and predicts a number of values as an output.

All these four cases, as shown in Table \ref{cases}, were trained on Bagged Trees, GPR, ARIMA, and LSTM in order to get predictions for the test data. Both the datasets, raw and smoothed, were used one by one to compare the predicted results.

\begin{table}[!h] 

\centering 

\caption{Number of Input and Output Parameters for all the cases} 

\resizebox{\columnwidth}{!}{ 

\begin{tabular}{|l|l|c|c|c|} 

\hline 

\multicolumn{1}{|c|}{\textbf{Cases}} & \textbf{Description} &\textbf{Features} & \textbf{\begin{tabular}[c]{@{}c@{}}Input\\ Months\end{tabular}} & \textbf{\begin{tabular}[c]{@{}c@{}}Output\\ Months\end{tabular}} \\ \hline 

Case 1& One month prediction with price data                                                                            & 1                 & 12                                                              & 1                                                                \\ 

Case 2& One month prediction with all features                                                                            & 6                 & 12                                                              & 1                                                                \\ 

Case 3& One year prediction with price data                                                                            & 1                 & 24                                                              & 12                                                               \\ 

Case 4& One year prediction with all features                                                                            & 6                 & 24                                                              & 12                                                               \\ \hline 

\end{tabular} 

} 

\label{cases} 

\end{table}

\section{Results}

Figure \ref{fig_orig} shows that the original pattern of prices increased over a period from 1991 to 2018 in the Faisalabad District of Pakistan.

The smoothing function extracts the right pattern of the data, which is more comfortable for the network to predict, thus improves the network's efficiency. All the networks are trained for both raw and smoothed data. The smoothed data showed better results compared to the raw data as it helps the networks to learn the trend better. For the validation of the network, we used RMSE as an error calculating term. Table \ref{case1}, \ref{case2}, \ref{case3}, and \ref{case4} shows the results obtained after training and testing the models mentioned above for each case.


\begin{table}[!h] 

\caption{Results for Case 1: One month prediction with price data} 

\centering 

\resizebox{\columnwidth}{!}{
\begin{tabular}{|c|c|c|c|c|c|c|}
\hline
\multirow{2}{*}{\textbf{Model}}                         & \multicolumn{2}{c|}{\textbf{\begin{tabular}[c]{@{}c@{}}Faisalabad\\ (RMSE)\end{tabular}}} & \multicolumn{2}{c|}{\textbf{\begin{tabular}[c]{@{}c@{}}Gujranwala\\ (RMSE)\end{tabular}}} & \multicolumn{2}{c|}{\textbf{\begin{tabular}[c]{@{}c@{}}Multan\\ (RMSE)\end{tabular}}} \\ \cline{2-7} 
                                                        & Raw                                         & Smooth                                      & Raw                                         & Smooth                                      & Raw                                       & Smooth                                    \\ \hline
\begin{tabular}[c]{@{}c@{}}Bagged \\ Trees\end{tabular} & 166.28                                      & 130.36                                      & 174.23                                      & 120.46                                      & 159.31                                    & 140.34                                    \\
GPR                                                     & 236.62                                      & 189.75                                      & 184.70                                      & 120.35                                      & 289.54                                    & 140.46                                    \\
ARIMA                                                   & 257.63                                      & 203.83                                      & 275.24                                      & 189.46                                      & 220.43                                    & 194.24                                    \\
LSTM                                                    & 159.19                                      & 98.28                                       & 160.21                                      & 96.13                                       & 164.23                                    & 104.57                                    \\ \hline
\end{tabular}
}
\label{case1}
\end{table}


\begin{table}[!h] 

\caption{Results for Case 2: One month prediction with all features} 

\centering 

\resizebox{\columnwidth}{!}{
\begin{tabular}{|c|c|c|c|c|c|c|}
\hline
\multirow{2}{*}{\textbf{Model}}                         & \multicolumn{2}{c|}{\textbf{\begin{tabular}[c]{@{}c@{}}Faisalabad\\ (RMSE)\end{tabular}}} & \multicolumn{2}{c|}{\textbf{\begin{tabular}[c]{@{}c@{}}Gujranwala\\ (RMSE)\end{tabular}}} & \multicolumn{2}{c|}{\textbf{\begin{tabular}[c]{@{}c@{}}Multan\\ (RMSE)\end{tabular}}} \\ \cline{2-7} 
                                                        & Raw                                         & Smooth                                      & Raw                                         & Smooth                                      & Raw                                       & Smooth                                    \\ \hline
\begin{tabular}[c]{@{}c@{}}Bagged \\ Trees\end{tabular} & 170.35                                      & 120.29                                      & 189.34                                      & 113.87                                      & 238.36                                    & 129.65                                    \\
GPR                                                     & 201.32                                      & 140.37                                      & 240.32                                      & 129.49                                      & 260.70                                    & 150.43                                    \\
ARIMA                                                   & 240.56                                      & 189.45                                      & 256.94                                      & 153.93                                      & 270.86                                    & 196.43                                    \\
LSTM                                                    & 136.34                                      & 78.45                                       & 149.24                                      & 71.62                                       & 170.23                                    & 82.33                                     \\ \hline
\end{tabular} 
}
\label{case2}
\end{table}


\begin{table}[h!] 

\caption{Results for Case 3: One year prediction with price data} 

\centering 

\resizebox{\columnwidth}{!}{
\begin{tabular}{|c|c|c|c|c|c|c|}
\hline
\multirow{2}{*}{\textbf{Model}}                         & \multicolumn{2}{c|}{\textbf{\begin{tabular}[c]{@{}c@{}}Faisalabad\\ (RMSE)\end{tabular}}} & \multicolumn{2}{c|}{\textbf{\begin{tabular}[c]{@{}c@{}}Gujranwala\\ (RMSE)\end{tabular}}} & \multicolumn{2}{c|}{\textbf{\begin{tabular}[c]{@{}c@{}}Multan\\ (RMSE)\end{tabular}}} \\ \cline{2-7} 
                                                        & Raw                                         & Smooth                                      & Raw                                         & Smooth                                      & Raw                                       & Smooth                                    \\ \hline
\begin{tabular}[c]{@{}c@{}}Bagged \\ Trees\end{tabular} & 259.34                                      & 242.54                                      & 257.45                                      & 220.64                                      & 284.76                                    & 229.35                                    \\
GPR                                                     & 286.78                                      & 230.74                                      & 314.58                                      & 246.53                                      & 298.34                                    & 236.69                                    \\
ARIMA                                                   & 268.56                                      & 220.34                                      & 270.47                                      & 257.63                                      & 268.80                                    & 243.84                                    \\
LSTM                                                    & 186.74                                      & 140.23                                      & 179.44                                      & 137.25                                      & 168.28                                    & 132.68                                    \\ \hline
\end{tabular}
}
\label{case3}
\end{table}


\begin{table}[!h] 

\caption{Results for Case 4: One year prediction with all features} 

\centering 

\resizebox{\columnwidth}{!}{
\begin{tabular}{|c|c|c|c|c|c|c|}
\hline
\multirow{2}{*}{\textbf{Model}}                         & \multicolumn{2}{c|}{\textbf{\begin{tabular}[c]{@{}c@{}}Faisalabad\\ (RMSE)\end{tabular}}} & \multicolumn{2}{c|}{\textbf{\begin{tabular}[c]{@{}c@{}}Gujranwala\\ (RMSE)\end{tabular}}} & \multicolumn{2}{c|}{\textbf{\begin{tabular}[c]{@{}c@{}}Multan\\ (RMSE)\end{tabular}}} \\ \cline{2-7} 
                                                        & Raw                                         & Smooth                                      & Raw                                         & Smooth                                      & Raw                                       & Smooth                                    \\ \hline
\begin{tabular}[c]{@{}c@{}}Bagged \\ Trees\end{tabular} & 205.34                                      & 187.69                                      & 213.44                                      & 170.32                                      & 234.37                                    & 1787.65                                   \\
GPR                                                     & 203.23                                      & 174.87                                      & 224.43                                      & 160.93                                      & 230.18                                    & 168.91                                    \\
ARIMA                                                   & 283.34                                      & 193.24                                      & 260.76                                      & 184.78                                      & 274.52                                    & 197.32                                    \\
LSTM                                                    & 185.73                                      & 146.83                                      & 204.61                                      & 150.72                                      & 172.13                                    & 148.21                                    \\ \hline
\end{tabular}
}
\label{case4}
\end{table}

\section{Comparison with related work}

The proposed methodology presented in this study to predict the prices of wheat in Pakistan is compared with the approaches used in similar previous works. Dataset of annual granularity was used in previous researches in predict the next year price value of wheat. We can see from Figure \ref{fig_orig} that the prices vary a lot within a single year and if we are able to predict the value of next month instead of singular value for an year, it would be more useful. Other than that in this study we focused on prices of wheat at district level, the prices vary from district to district because of multiple reasons like transport, weather and population. So if we are able to predict the price of wheat for a mandi in a specific district it would also helps the authorities to keep track on the price variation within districts.  Table \ref{comp} compares the techniques, dataset, and their prediction errors with our proposed method (LSTM) to predict prices for a single month as well as for a whole year consisting of 12 months. The single month prediction is much better as than the whole year one as it is predicting lesser amount of data.


\begin{table*}[b] 

\caption{Comparison with previous studies} 

\centering 

\begin{tabularx}{0.6955\textwidth}{|l|l|l|l|l|r|} 

\hline 

\multicolumn{1}{|c|}{\textbf{Sr.}} & \multicolumn{1}{c|}{\textbf{Study}} & \multicolumn{1}{c|}{\textbf{Method}} & \multicolumn{1}{c|}{\textbf{Data}} & \multicolumn{1}{c|}{\textbf{Prediction}} & \multicolumn{1}{c|}{\textbf{\begin{tabular}[c]{@{}c@{}}Error/\\ Performance\\ (MAPE)\end{tabular}}} \\ \hline 

1 & \begin{tabular}[c]{@{}l@{}}Rani and Raza\\ \cite{e1}\end{tabular} & \begin{tabular}[c]{@{}l@{}}Double \\ exponential \\ smoothing\end{tabular} & \begin{tabular}[c]{@{}l@{}}Yearly average price\\ of gram in Pakistan\\ 1975 - 2010\end{tabular} & Next year forecast & 28.19 \\ 

2 & \begin{tabular}[c]{@{}l@{}}Saqib Shakeel\\ et al. \cite{c1}\end{tabular} & S-Curve fitting & \begin{tabular}[c]{@{}l@{}}Yearly average price \\ of wheat in Paksitan\\ 1975 - 2013\end{tabular} & Next year forecast & 12.60 \\ 


4 & \begin{tabular}[c]{@{}l@{}}Our proposed\\ method\end{tabular} & LSTM & \begin{tabular}[c]{@{}l@{}}Monthly wheat prices\\ with useful features \\ 1991 - 2018\\ \end{tabular} & \begin{tabular}[c]{@{}l@{}}Next month\\ forecast\\ \\ Next 12 months\\ forecast\end{tabular} & \begin{tabular}[c]{@{}r@{}}0.71\\ \\ \\ 1.27\end{tabular} \\ \hline 

\end{tabularx} 

\label{comp} 

\end{table*} 



\section{Conclusions}

In this study, we have proposed an advance and  efficient neural network system to forecast the prices of wheat in districts of Pakistan using LSTM and provided its comparison with other popular machine learning time series techniques. We also used the smoothness algorithm on the raw data. This smoothness algorithm helps the models to predict the values more efficiently compared to the raw data used. We managed to get state of art results for Pakistan wheat prices using the LSTM model as compared to the previous techniques used for the same problem.  

The proposed model offers systems to predict prices of wheat for a single month as well as for 12 months (whole year) in future at a time. Our proposed system shows significant improvement in terms of prediction errors compared to the already existing works in this area. Therefore, this study can prove to be crucial in Pakistan's agricultural economic studies, especially in making policies, as it predicts the prices of the wheat by thinking ahead in time with more acceptable accuracy providing more confidence in drawn insights.


\begin{IEEEbiography}[{\includegraphics[width=1in,height=1.25in,clip,keepaspectratio]{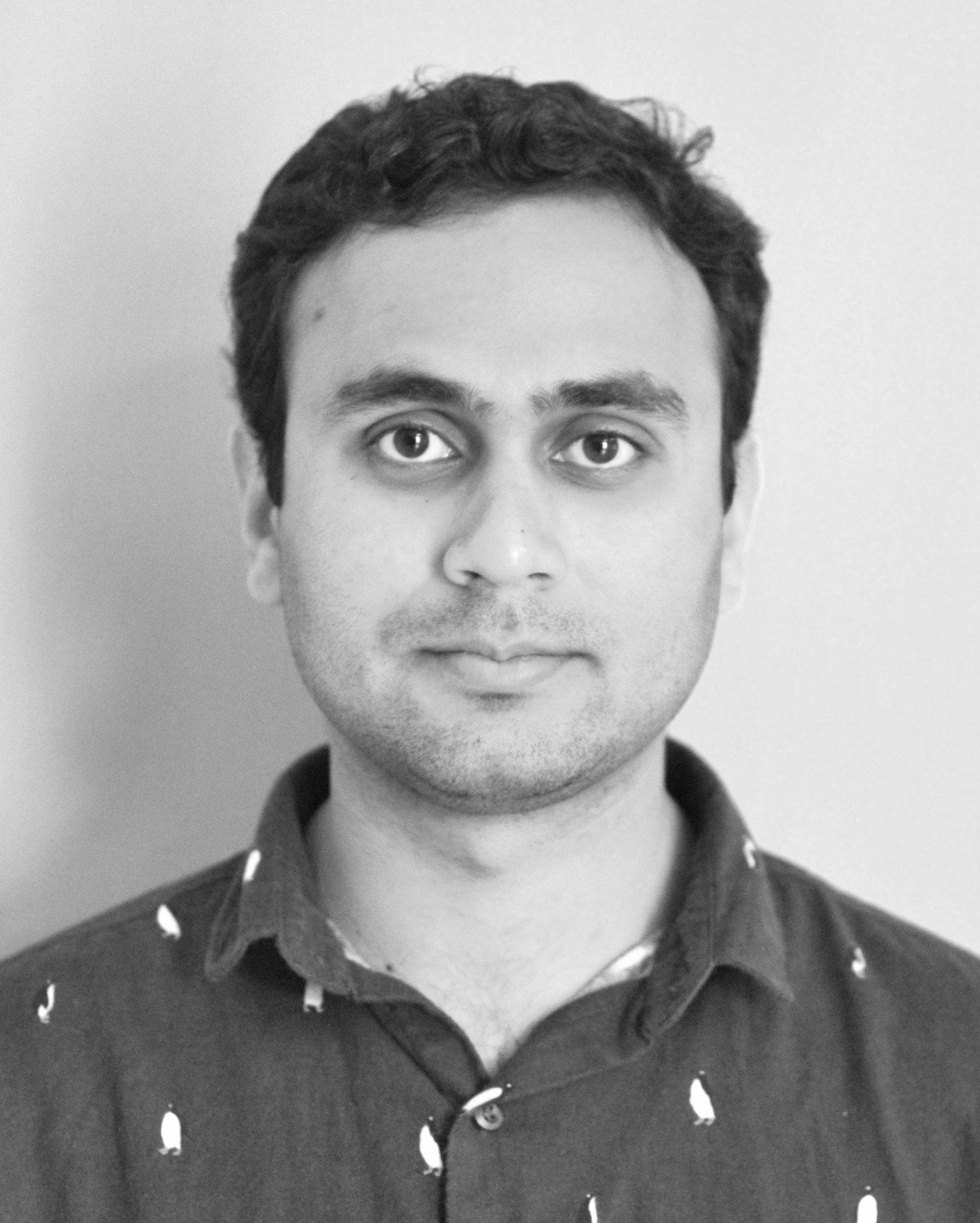}}]{Ahmed Rasheed} earned his Bachelor's degree in Electrical Engineering, majoring in Electrical Power, from Air University Islamabad, Pakistan. He is currently an Electrical Engineering Master's student majoring in Digital System and Signal Processing, at School of Electrical Engineering and Computer Science (SEECS) at the National University of Sciences and Technology (NUST) Islamabad, Pakistan. He is also serving as a Research Assistant at the laboratory of Adaptive Signal Processing (ASP) at SEECS, specializing in handling machine learning, deep learning, time series prediction and biomedical image processing tasks.

\end{IEEEbiography}

\begin{IEEEbiography}[{\includegraphics[width=1in,height=1.25in,clip,keepaspectratio]{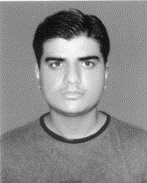}}]{Muhammad Shahzad Younis} received the bachelor's degree from National University of Sciences and Technology, Islamabad, Pakistan, in 2002, the master's degree from the University of Engineering and Technology, Taxila, Pakistan, in 2005, and the PhD degree from University Technology PETRONAS, Perak, Malaysia in 2009, respectively. Before joining National University of Sciences and Technology (NUST), he was Assistant Manager at a research and development organization named AERO where he worked on different signal processing and embedded system design applications. He is currently working as an Assistant professor in the Department of Electrical Engineering in School of Electrical Engineering and Computer Science (SEECS)-NUST. He has published more than 25 papers in domestic and international journals and conferences. His research interests include Statistical Signal Processing, Adaptive Filters, Convex Optimization Biomedical signal processing, wireless communication modelling and digital signal processing.  

\end{IEEEbiography}

\begin{IEEEbiography}[{\includegraphics[width=1in,height=1.25in,clip,keepaspectratio]{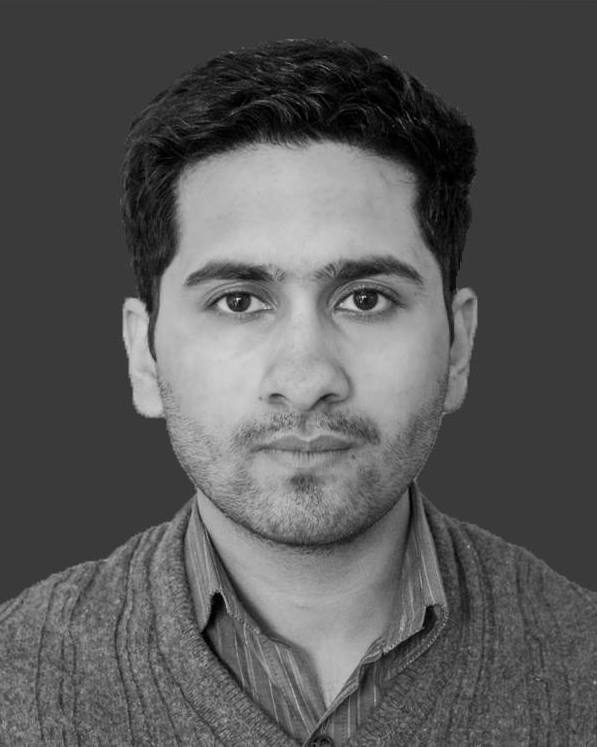}}]{Farooq Ahmad} earned his Bachelor's degree in Electrical Engineering(Electronics) from Air University Islamabad, Pakistan. He is currently an Electrical Engineering Master's student majoring in Power and Control Systems, at School of Electrical Engineering and Computer Science (SEECS), National University of Sciences and Technology (NUST) Islamabad, Pakistan. He is also serving as a Project Engineer at the laboratory of Adaptive Signal Processing (ASP) at SEECS. His research interest is in in IoT and data-driven control. 

\end{IEEEbiography}

\begin{IEEEbiography}[{\includegraphics[width=1in,height=1.25in,clip,keepaspectratio]{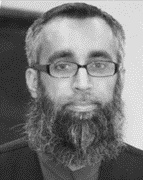}}]{Junaid Qadir} (SM? 14) completed his BS in Electrical Engineering from UET, Lahore, Pakistan and his PhD from University of New South Wales, Australia in 2008. He is currently an Associate Professor at the Information Technology University (ITU)-Punjab, Lahore, Pakistan. He is the Director of the IHSAN Lab at ITU that focuses on deploying ICT for development and is engaged in systems and networking research. Prior to joining ITU, he was an Assistant Professor at the School of Electrical Engineering and Computer Sciences (SEECS), National University of Sciences and Technology (NUST), Pakistan. At SEECS, he directed the Cognet Lab at SEECS that focused on cognitive networking and the application of computational intelligence techniques in networking. He has been awarded the highest national teaching award in Pakistan?the higher education commission's (HEC) best university teacher award?for the year 2012-2013. He has been nominated for this award twice (2011, and 2012-2013). His research interests include the application of algorithmic, machine learning, and optimization techniques in networks. In particular, he is interested in the broad areas of wireless networks, cognitive networking, software-defined networks, and cloud computing. He is a regular reviewer for a number of journals and has served in the program committee of a number of international conferences. He serves as an Associate Editor for IEEE Access, IEEE Communication Magazine, and Springer Nature Big Data Analytics. He was the lead guest editor for the special issue "Artificial Intelligence Enabled Networking" in IEEE Access and the feature topic "Wireless Technologies for Development" in IEEE Communications Magazine. He is a member of ACM, and a senior member of IEEE. 

\end{IEEEbiography}

\begin{IEEEbiography}[{\includegraphics[width=1in,height=1.25in,clip,keepaspectratio]{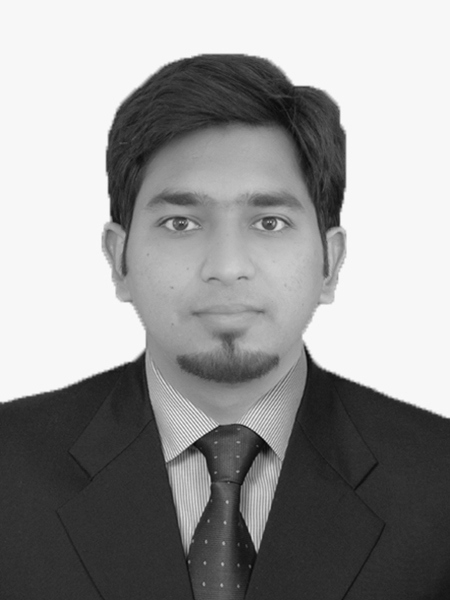}}]{Muhammad Kashif} received his Bachelor in Electronic Engineering from Ghulam Ishaq Khan Institute of Engineering, Sciences and Technology (GIKI), Swabi, Pakistan in 2016. He has worked as Trainee Engineer in Thermovision (PVT) ltd from 2016-17. He is currently enrolled in MS Electrical Engineering (Power and Control Systems) Program 2017 in NUST, Islamabad, Pakistan. He is working as Research Assistant in ASP Lab, SEECS, NUST. His research interest is in power system monitoring and analysis, development of smart sensors.

\end{IEEEbiography}

\EOD

\end{document}